\documentclass[letterpaper]{article} 
\pdfoutput=1  
\usepackage[preprint]{aaai2027}
\usepackage[hyphens]{url}  
\usepackage{graphicx} 
\usepackage{natbib}  
\usepackage{caption} 
\usepackage{amsmath,amssymb}
\usepackage{booktabs}
\usepackage{xcolor}

\graphicspath{{Figures/}}

\title{When Is Noise Response Universal?\\Tokenization as the Hidden Variable in Language Models}

\author{Yefan Tao, Gerald Friedland, Luyang Kong}
\affiliations{Amazon\\ \{tayefan, gfriedla, luyankon\}@amazon.com}

\begin{document}
\maketitle

\begin{abstract}
The performance of textual neural models often degrades when their inputs are corrupted by noise such as typos, OCR errors, or dropped words. We study the degradation rate across neural models, both sentence embeddings and decoder-only LLMs, and find that how consistent it is depends on the scale of the noise: under word-level noise, models with very different architectures decline along nearly the same curve, while under character-level noise they separate. We further identify the determining factor to be the training objective, not the architecture: eight encoders spanning six pretraining paradigms are scattered initially, and collapse onto a common curve after a short contrastive training recipe. We trace the word/character split to tokenization: a single character edit forces the tokenizer to re-segment the surrounding word, disturbing the token sequence far more than dropping a whole word does. This finding and its underlying mechanism provide a practical means to predict a model's robustness to noise without any noisy evaluation, and to install robustness at a chosen noise scale through noise-augmented training.
\end{abstract}

\section{Introduction}
\label{sec:intro}
Deployed NLP systems routinely encounter noisy inputs: OCR pipelines misread characters (l$\to$1, rn$\to$m), speech-to-text systems introduce phonetic substitutions, and users produce typos in every text field from search queries to chatbot prompts~\citep{morris2020textattack}. Understanding how models degrade under such noise is critical for reliability, but practitioners currently lack principled tools to predict robustness without exhaustive per-model evaluation. We show that such prediction is often possible, because architecturally different models frequently trace the same normalized degradation curve. We call this universality, in analogy to statistical physics, where microscopically different systems exhibit identical macroscopic behavior, captured by shared scaling functions and exponents once each system is rescaled by its own characteristic scale, determined by a few coarse features such as symmetry and dimensionality~\citep{kadanoff1966scaling,wilson1971renormalization}.

This universality is uneven across noise types. Sentence-embedding models degrade near-identically under word-level noise (coefficient of variation, CV, of 2\% across nine models) but visibly less so under character-level noise (CV 14\%). The split is largely an artifact of tokenization: every model sees input only after it is split into tokens, and the two noise types are not equally violent there. Dropping a word removes its tokens and leaves the rest intact; substituting one character forces the tokenizer to re-segment the surrounding word, shattering a token into several unrelated subwords. At a matched corruption rate this makes character noise induce 4--10$\times$ more token-edit distance than word dropout, and because the amount of shatter is tokenizer-specific, models with different vocabularies spread apart under it while their word-noise responses stay together.

This tokenization mechanism explains why models sharing a tokenizer diverge much less under character noise than those with different tokenizers, and why a character-level backbone, which cannot shatter, stays character-robust even under an identical objective. Token-edit distance is a strong proximate driver but not a sufficient statistic, since at matched token distance removing a content word still perturbs a representation more than respelling one. The mechanism is also practical: within a regime, universality lets a model's clean accuracy forecast its degradation curve without any noisy evaluation, and noise-augmented training installs robustness at a chosen token-perturbation scale, improving accuracy under noise by up to 19\% while preserving clean performance.

\paragraph{Contributions.}
\begin{enumerate}
\item \textbf{A universality split by noise scale.} Across sentence embeddings and 18 decoder-only LLMs, word-noise degradation curves are near-identical (cross-model CV $\sim$2\%) while character-noise curves separate. We judge this the way physics defines a universality class, by a shared decay exponent rather than a hand-set threshold, and corroborate it with two fitting-free measures (cross-model CV and the intraclass correlation, ICC).

\item \textbf{The split is caused by the training objective, not the architecture.} Eight encoders spanning six pretraining paradigms are scattered before a shared contrastive objective and collapse onto one curve after it.
\item \textbf{A token-perturbation mechanism.} A character edit forces re-tokenization that perturbs the token sequence 4--10$\times$ more than a word edit, in a tokenizer-specific way; a single corruption knob that tunes this perturbation reproduces the whole universal-to-divergent spectrum. Token perturbation is a strong mediator, not the sole variable.
\item \textbf{Practical consequences.} Within the universal regime a model's degradation curve is predictable from clean accuracy alone (3.5\% leave-one-out error), and noise-augmented training installs robustness at a chosen scale; but such post-hoc robustness does not create universality, which stems from the shared training-time constraint.
\end{enumerate}

\section{Related Work}
\label{sec:related}

\paragraph{Robustness to noisy text.}
A line of work documents how NLP models degrade when inputs contain typos, OCR errors, or speech-to-text noise. \citet{belinkov2018synthetic} show that character-level neural machine translation systems are highly sensitive to natural and synthetic noise (scrambling, swap, keyboard typos), with BLEU dropping by tens of points. \citet{pruthi2019combating} introduce a word-recognition defense against character-level adversarial misspellings, and \citet{moradi2021evaluating} extend this analysis to BERT-class models across tasks. In production, spelling correction and text normalization remain the dominant approach. These works characterize individual models' degradation; we ask the orthogonal question of whether different models degrade according to a common law.

\paragraph{Adversarial robustness.}
A separate body of work crafts minimal perturbations to fool a target model, including paraphrase attacks~\citep{jia2017adversarial}, unified attack frameworks~\citep{morris2020textattack}, and benchmark suites for adversarial evaluation~\citep{wang2021adversarial}. For LLMs, recent attention has focused on prompt sensitivity~\citep{lu2022fantastically} and adversarial jailbreaking~\citep{zou2023universal}. Adversarial perturbations are crafted against a specific target model and behave differently across models by design. We instead study how models respond to random, naturally occurring noise, and find that whether this response is shared across models depends on the noise scale: it is shared under word-level noise but not under character-level noise, a split that tracks how each model's tokenizer reacts to the corruption.

\paragraph{Contrastive sentence representations.}
Sentence embeddings have converged on a contrastive paradigm: Sentence-BERT~\citep{reimers2019sentence}, SimCSE~\citep{gao2021simcse}, E5~\citep{wang2022text}, and BGE~\citep{xiao2024c} all learn from pairwise similarity signals, with later models adopting InfoNCE-style objectives explicitly. \citet{wang2020understanding} characterize the geometry these objectives induce in terms of alignment and uniformity on the hypersphere. We build on this characterization to argue that the shared training objective itself, not the architecture, is what gives rise to a shared noise-response curve across these models.

\paragraph{Universality and convergence in neural networks.}
In statistical physics, universality describes how microscopically different systems share macroscopic behavior~\citep{kadanoff1966scaling}. \citet{bahri2020statistical} survey connections of these ideas to neural networks. A separate thread documents universality of scale rather than architecture, including neural scaling laws~\citep{kaplan2020scaling} and emergent capabilities~\citep{wei2022emergent}. 
The closest line of work to us is on representational convergence across architectures. \citet{kornblith2019similarity} introduce centered kernel alignment (CKA) and show that hidden representations of differently-initialized or differently-architected networks become highly aligned when trained on the same data. The Platonic Representation Hypothesis~\citep{huh2024platonic} extends this across modalities, arguing that vision and language models converge to a shared distance geometry as they scale. These works share our motivation of asking what is invariant across architecturally distinct models, but study a different invariant: static geometry, in the form of pairwise distances or kernel similarity computed on clean inputs. Our universality is dynamic, capturing how a model's output changes as noise is injected.

\section{Setup}
\label{sec:setup}

We study two families of neural text models, sentence-embedding models and decoder-only LLMs, and measure how a model's output changes as its input is corrupted. A sentence-embedding model emits a vector, whose quality we read out as a similarity-based task score (primarily STS-B Spearman correlation). A decoder-only LLM emits a next-token distribution, whose shift we read out as the Jensen--Shannon (JS) divergence between the clean and noisy distributions. The noise types and model set are shared across both regimes.

\paragraph{Why these metrics.}
Both choices target the layer where the training objective acts. Contrastive embedding models are trained to map similar inputs to nearby vectors, so a similarity-based downstream task (e.g., STS-B Spearman correlation) directly probes what the loss optimizes. LLMs are trained to match a target next-token distribution, so the natural noise-response signal is a divergence between distributions, not a downstream accuracy that also reflects task structure. JS divergence is symmetric, bounded in $[0, \log 2]$, and well-defined when distributions have disjoint support. We report the retained agreement $1-\text{JS}$, which decays from one as noise perturbs the output, matching the embedding score.

\paragraph{Noise types.} We apply four input-level perturbations, each parameterized by a single rate $\varepsilon \in [0, 0.5]$, the fraction of units (characters or words) corrupted, evaluated at 13 levels: \textbf{character noise} (each character replaced by a uniform random one with probability $\varepsilon$), \textbf{word dropout} (each whitespace-separated word removed with probability $\varepsilon$), \textbf{keyboard typo} (QWERTY-adjacent substitutions), and \textbf{OCR error} (common confusion pairs such as l$\leftrightarrow$1, rn$\leftrightarrow$m). Each curve averages over 1{,}000+ inputs, so a single noise seed suffices for the main measurements (re-seeding shifts the aggregate by $<$0.3\% relative); the decay-exponent and ICC criteria are seed-insensitive by construction, and the sensitive before/after training comparison uses ten seeds per level.

\paragraph{Embedding evaluation.} We evaluate embedding models on four tasks spanning their uses: \textbf{STS-B}~\citep{cer2017semeval} (Spearman $\rho$ of cosine vs.\ human similarity; the primary, main-text readout), \textbf{20 Newsgroups}~\citep{pedregosa2011scikit} (clustering silhouette), \textbf{AG News}~\citep{zhang2015character} (nearest-centroid accuracy), and \textbf{SciFact}~\citep{thakur2021beir} (nDCG@10 retrieval, query noised); the latter three agree with STS-B and appear in Appendix~\ref{app:cross_dataset}. Each task yields a quality score $\rho(\varepsilon)$ that we fit with the normalized degradation curve
\begin{equation}
\label{eq:master}
\frac{\rho(\varepsilon)}{\rho(0)} = A\, e^{-k\varepsilon} + (1-A),
\end{equation}
where $k$ is the decay rate and $1-A$ the noise-invariant offset (Appendix~\ref{app:stretched} examines a more flexible stretched-exponential form: it is an adequate memoryless fit under character noise, while word noise shows mild cooperativity).

\paragraph{LLM evaluation.} For each input we measure the mean Jensen--Shannon divergence $\text{JS}(\varepsilon)$ between the clean and noisy next-token distributions, over 1000 sentences pooled from five domains: \textbf{AG News}~\citep{zhang2015character}, \textbf{Wikitext-103}~\citep{merity2016pointer}, \textbf{SST-2}~\citep{socher2013recursive}, \textbf{SQuAD}~\citep{rajpurkar2016squad}, \textbf{SNLI}~\citep{bowman2015large}. We then track the retained agreement $1-\text{JS}(\varepsilon)$, which starts at one and decays, fit with the same form (Eq.~\ref{eq:master}),
\begin{equation}
\label{eq:llm_master}
1 - \text{JS}(\varepsilon) = A\, e^{-k \varepsilon} + (1 - A).
\end{equation}
Using $1-\text{JS}$ anchors the curve at one (since $\text{JS}(0)=0$); the choice does not affect any verdict below, which is unchanged under the raw $\text{JS}/\text{saturation}$ curve (Appendix~\ref{app:cross_dataset}). Where a coarser readout helps (the intervention experiments) we also report \textbf{top-1 match}, the fraction of inputs whose argmax token is unchanged.

\paragraph{Models.} Our model set spans the axes universality should be invariant to: architecture, scale, pooling, and training lineage (full specifications in Appendix~\ref{app:models}). \textbf{Sentence embeddings (9, contrastively trained):} all-MiniLM-L6-v2, all-MiniLM-L12-v2, all-mpnet-base-v2, bge-base-en-v1.5, bge-large-en-v1.5, e5-base-v2, e5-large-v2, gtr-t5-base, and nomic-embed-text-v1.5, spanning 23M--335M parameters, four backbones (MiniLM, MPNet, BERT, T5 encoder), three embedding dimensions, and both mean and CLS pooling. \textbf{Non-sentence-trained baselines (4):} bert-base-uncased, bert-large-uncased, roberta-base, and e5-mistral-7b, which share backbones and tokenizers with the first group but never receive the contrastive objective, isolating its effect. \textbf{Decoder-only LLMs (18, 9 families):} GPT2-XL; Qwen2.5-\{1.5B, 7B\} and Qwen2.5-\{3B, 7B, 14B, 32B, 72B\}-Instruct; Qwen3-\{1.7B, 8B\}; Phi-2; Phi-3-mini; Llama-2-7B; Llama-2-7B-Chat; Llama-3-8B-Instruct; Mistral-7B-Instruct; DeepSeek-7B-Chat; and Yi-1.5-9B-Chat, covering a 48$\times$ scale range (1.5B--72B), nine families, and two generations (Qwen2.5 and Qwen3), so that any shared degradation cannot be attributed to a single lineage.

\paragraph{When is a response universal?}
We summarize cross-model agreement with the pointwise coefficient of variation (CV) of the normalized degradation curves. Let $c_m(\varepsilon) = q_m(\varepsilon)/q_m(0)$ be model $m$'s degradation curve normalized by its own clean value, where the quality $q$ is the task score for embeddings and the retained agreement $1-\text{JS}$ for LLMs. The CV is the across-model standard deviation divided by the mean at each noise level, averaged over $\varepsilon>0$:
\begin{equation}
\label{eq:cv}
\mathrm{CV} = \operatorname*{mean}_{\varepsilon>0}\;\frac{\operatorname{std}_m\, c_m(\varepsilon)}{\bigl|\operatorname{mean}_m\, c_m(\varepsilon)\bigr|}.
\end{equation}
We call a response universal when models share the same decay rate, mirroring how a universality class is defined in statistical physics by a shared exponent. We fit each model's normalized curve with a single exponential of decay rate $k_m$ and amplitude $A_m$ (the form is Eq.~\ref{eq:master} for embeddings and Eq.~\ref{eq:llm_master} for LLMs), attach a bootstrap confidence interval to each $k_m$, and call the models one universality class when those intervals overlap. This is cleanest for embeddings, where $A_m \approx 1$ (a pure exponential) leaves $k_m$ unambiguous; it is our primary criterion there.

Two further measures, both free of curve fitting, cross-check the exponent and extend to both architectures. The CV is dimensionless, so the ratio between two conditions' CVs is itself a unitless effect size that grows with disagreement. The ICC is the fraction of the per-input response variance attributable to model identity, bounded in $[0,1]$: near zero the response is model-independent, near one it is model-specific. We compute it at the per-input level (per-sentence for LLMs, per-pair cosine retention for embeddings) so that it does not grow with dataset size (Appendix~\ref{app:ruler}). We read universality from the agreement of these three measures. For LLMs the amplitude $A_m$ varies across models (shown in the LLM section), as physics expects amplitudes to vary across members of the same class, so there we compare the exponent after dividing $A_m$ out and lead with the CV and ICC.

\section{Universality of Embedding Models}
\label{sec:emb_results}

\subsection{Universal Under Word Noise, Not Character Noise}
\label{sec:emb_phenomenon}

We measure each model's normalized degradation curve and ask whether architecturally different models follow the same one, taking the three measures from the Setup in order of how directly they test universality.

\begin{figure}[htbp]
\centering
\includegraphics[width=\columnwidth]{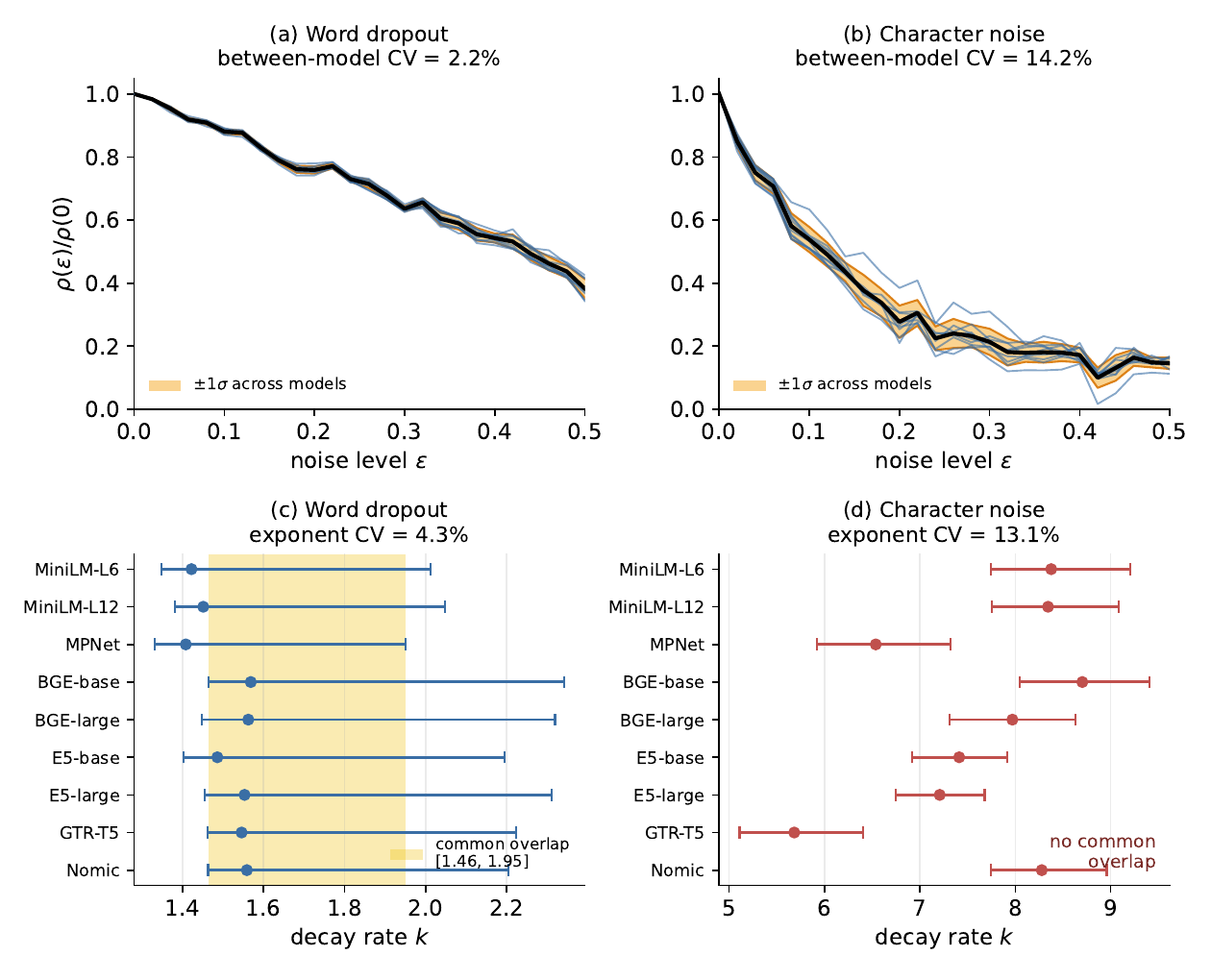}
\caption{The embedding word/character split (word left, character right). \textbf{Top:} the nine models' normalized curves; the amber $\pm1\sigma$ across-model band is thin under word noise (CV 2.2\%), six times wider under character noise (14.2\%). \textbf{Bottom:} per-model decay rate $k_m$ with bootstrap 95\% CI. Under word noise every interval crosses a common $k\approx1.5$ (shaded): one universality class; under character noise the rates spread with no common overlap ($A_m\approx1$, so $k_m$ is unambiguous).}
\label{fig:emb_split}
\end{figure}

\paragraph{Shared decay exponent (primary).}
\label{sec:emb_exponent}
We fit the single exponential of Eq.~\ref{eq:master} to each model and read off its decay rate $k_m$ with a bootstrap confidence interval; the models are one universality class when these intervals share a common overlap. For all nine the fitted amplitude is $A_m\approx1.00$ (a pure exponential), leaving $k_m$ unambiguous. Under word noise the nine rates are nearly identical, all between $1.41$ and $1.57$, and every interval contains $k\approx1.5$: one decay rate, a single universality class. Under character noise the rates spread from $5.68$ to $8.70$, with no value inside all the intervals, so the models fall into different classes (Figure~\ref{fig:emb_split}c,d). The amplitude does not enter this comparison, which makes it our cleanest statement of the split.

\paragraph{Cross-model CV and ICC agree.}
\label{sec:emb_icc}
The same split is visible directly in the curves (Figure~\ref{fig:emb_split}a,b): under word dropout the nine models collapse onto a common curve, under character noise they spread apart. Two measures that need no curve fit make this quantitative. The cross-model CV is \textbf{2.2\%} under word dropout and \textbf{14.2\%} under character substitution, six times larger. The ICC measures how much of the variance in a model's noise response is explained by which model it is, rather than by the input: model identity accounts for only \textbf{14\%} of the variance under word noise but \textbf{60\%} under character noise (jackknife $0.14{\pm}0.01$ vs $0.60{\pm}0.02$). In other words, under word noise the models behave almost interchangeably, while under character noise each responds in its own way. Detailed per-noise-type CVs are in Appendix~\ref{app:supp}; the split survives before any normalization (Appendix~\ref{app:normalization}) and across three other embedding tasks (Appendix~\ref{app:cross_dataset}).

\subsection{The Objective Installs Universality}
\label{sec:causal}

What causes this universality? We find that it is driven by the training objective, not the architecture. To show this, we take eight encoders spanning six pretraining paradigms, BERT, RoBERTa, and DistilBERT (masked-LM), ALBERT (parameter sharing), ELECTRA (replaced-token detection), DeBERTa (disentangled attention), XLNet (permutation LM), and the T5 encoder (encoder--decoder), and measure their word-dropout curves as pretrained. Their decay exponents are all over the map, $k_m$ from $2.1$ to $9.7$ (CV of the exponent \textbf{71\%}; inter-model CV \textbf{14.3\%}): off-the-shelf models of different architectures are not one universality class to begin with. We then contrastively train each from its pretrained checkpoint under one identical recipe (InfoNCE, $\tau{=}0.05$, 50K SNLI entailment pairs, 5K steps) and re-measure. Despite the incompatible architectures, their exponents converge to a single value, $k_m\in[1.41,1.57]$ (CV of the exponent \textbf{3.2\%}; inter-model CV from 14.3\% to \textbf{2.8\%}), the same $k\approx1.5$ found for the nine production models earlier in this section (Figure~\ref{fig:arch_univ}; per-model numbers in Appendix~\ref{app:arch}). That two unrelated sets of models land on the same exponent is itself evidence the class is real. This before/after contrast is the causal statement: holding the noise and evaluation fixed and varying only whether the objective has been applied, the objective, not the architecture or the starting weights, is what moves a model onto the curve. The converse holds too: four models that never receive the sentence-level objective, BERT and RoBERTa (masked-LM) and E5-Mistral (a decoder backbone), stay off the common curve, with inter-model CV above 20\%.

Character noise confirms this while exposing the objective's limit. Training pulls the character curves closer too (CV 29.7\% to 13.0\%), but only partly: at 13.0\% the eight stay about as scattered as the untrained word-noise baseline, so they remain non-universal under character noise. The objective squeezes both scales, yet only word noise collapses to the near-2\% regime, because character noise is scrambled by each model's tokenizer before it reaches the shared geometry, so the common fixed point cannot enforce a common response there.

\begin{figure}[t]
\centering
\includegraphics[width=\columnwidth]{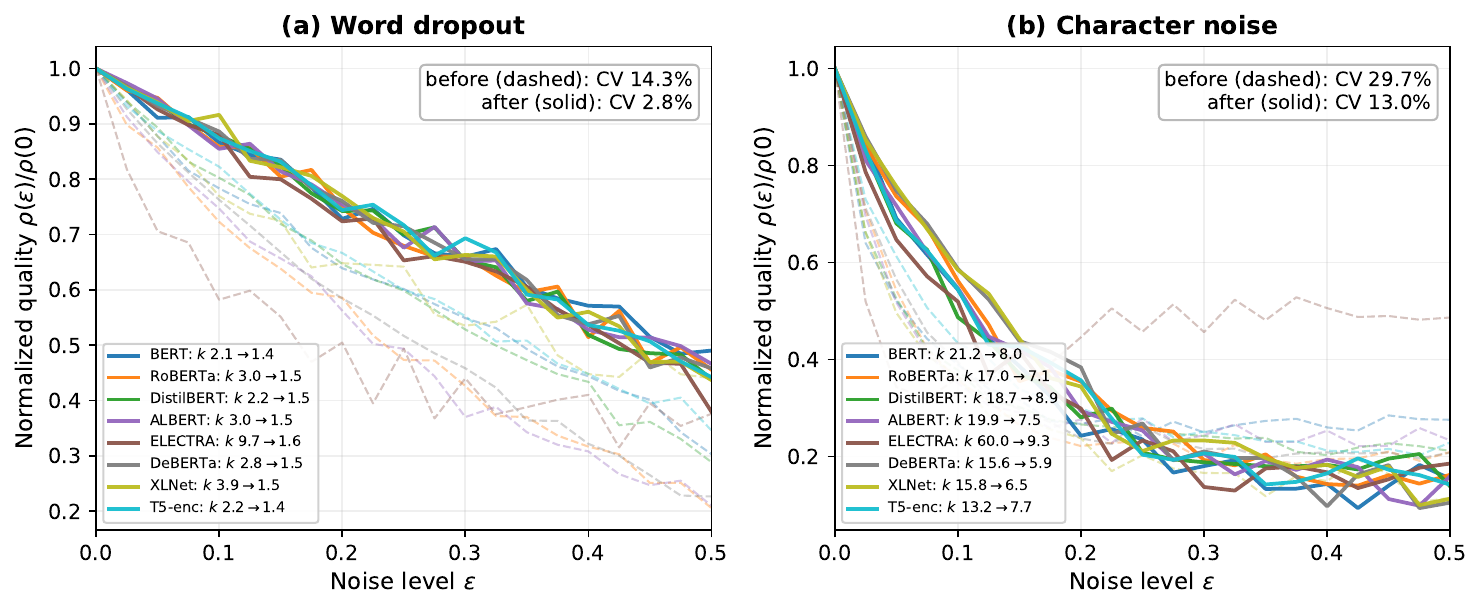}
\caption{Eight encoders with different pretraining paradigms, before contrastive training (dashed) and after an identical recipe (solid). Under word dropout (left) the scattered rates converge to $k\approx1.5$ (CV 14.3\%$\,\to\,$2.8\%); under character noise (right) training tightens them only partly (29.7\%$\,\to\,$13.0\%).}
\label{fig:arch_univ}
\end{figure}

\section{Universality of Decoder-Only LLMs}
\label{sec:llm_results}

\begin{figure}[t]
\centering
\includegraphics[width=0.9\columnwidth]{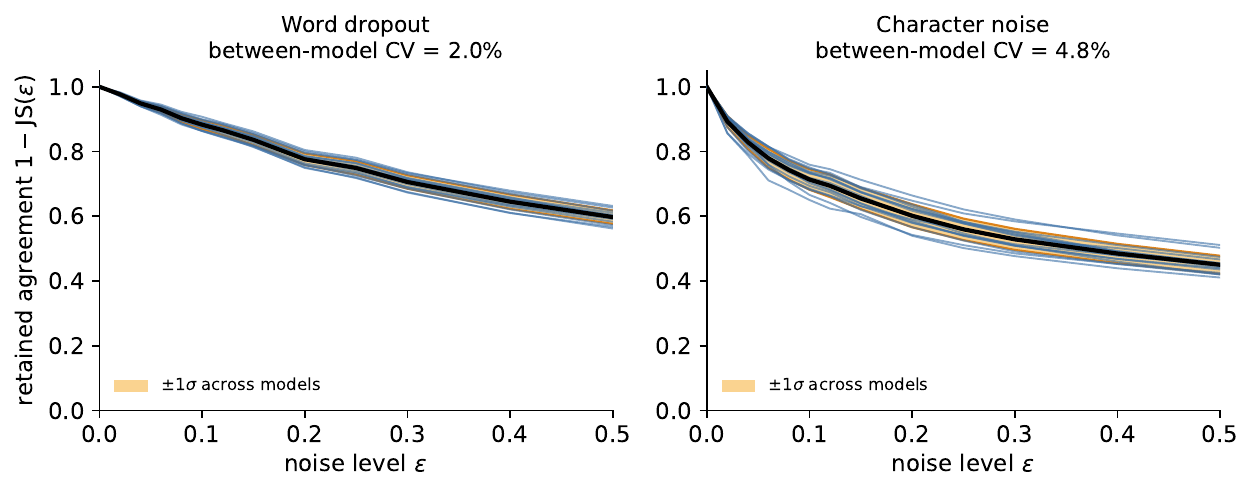}
\caption{Decoder-only LLMs show the same word/character split as embeddings. Each curve is the retained agreement $1-\text{JS}(\varepsilon)$; the amber shading is the $\pm1\sigma$ spread across models (centered on the cross-model mean, black), whose width is the cross-model CV. It is thin under word dropout (left, CV 2.0\%) and widens under character noise (right, CV 4.8\%). The two largest models (32B, 72B), run on a coarser grid, appear in Appendix~\ref{app:models} only.}
\label{fig:llm_univ}
\end{figure}

The natural readout of an LLM is its output distribution, so for each of 18 models across 9 families we measure the Jensen--Shannon divergence between clean and noisy next-token distributions and form the normalized degradation curve, exactly as on the embedding side. The same word/character split appears (Figure~\ref{fig:llm_univ}). The order of evidence reverses the embedding case: the LLM amplitude $A_m$ varies across models, so the fitted exponent is not read off as cleanly, and we lead with the fitting-free CV and ICC, reporting the amplitude-controlled exponent as support.

\paragraph{Cross-model CV.} Under word dropout the retained-agreement curves $1-\text{JS}(\varepsilon)$ collapse onto a common curve, with a normalized-curve CV of \textbf{2.0\%} (Figure~\ref{fig:llm_univ}, left), matching the embedding word-noise value of 2.2\% despite sharing nothing with embeddings but the training signal. Under character noise the spread rises to \textbf{4.8\%} (right). A bootstrap over models places the word CV at 2.0\% (95\% CI $[1.4, 2.3]$) and the character CV at 4.8\% ($[3.3, 5.8]$); the two intervals do not overlap.\footnote{Word removal renormalizes whitespace, so the clean JS reference must be computed on the same whitespace-normalized text; comparing against the raw sentence inflates the apparent spread. We measure against the matched baseline so every curve starts at $\mathrm{JS}=0$.} The split is the same as for embeddings, milder in magnitude but identical in ordering and cause.

\paragraph{Intraclass correlation.} The ICC, the fraction of the per-sentence noise-response variance attributable to which model produced it, needs no spread-versus-spread comparison and does not depend on evaluation size. Model identity accounts for only \textbf{0.7\%} of the word-noise variance but \textbf{4.6\%} of the character-noise variance, a roughly sixfold increase (a leave-one-model-out jackknife places the two cleanly apart, $0.007{\pm}0.002$ versus $0.046{\pm}0.012$). Both are small, but character noise is where model identity matters most. These absolute values are not comparable to the embedding ICC, whose per-pair cosine readout is on a different scale; what carries across both architectures is that the character-noise ICC far exceeds the word-noise one. The spread runs across all 9 families and the full 48$\times$ scale range, with per-model values in Appendix~\ref{app:models}.

\paragraph{Decay exponent.} The decay-exponent criterion that is so clean for embeddings needs an extra step here, for an informative reason. Fitting $A_m e^{-k_m\varepsilon}+(1-A_m)$ to the LLM curves returns an amplitude $A_m$ that varies from $0.53$ to $0.75$: a per-model prefactor, exactly the role a non-universal amplitude plays in a universality class where only the exponent is universal. The amplitude is not the universal quantity, and it does carry some scale dependence (larger models have a smaller word-noise $A_m$, Spearman $\rho{=}{-}0.66$, $p{=}0.01$; the character-noise $A_m$ is uncorrelated with size, $\rho{=}0.24$, $p{=}0.33$). Comparing exponents therefore requires dividing the amplitude out (saturation normalization), and the word exponents are then again far tighter than the character exponents ($5.3\%$ versus $15.2\%$ dispersion, joint-bootstrap 95\% CIs $[4.8,6.4]$ versus $[12.6,20.4]$, non-overlapping; Appendix~\ref{app:exponent}), the ordering the CV and ICC give. The embedding exponent (where $A\approx1$) carries the clean version of this argument.

\paragraph{The spread holds for realistic noise.} Three structured character-noise variants on a subset of 8 LLMs (keyboard typos, OCR confusions, adjacent-character swaps) all show the same elevated spread, with normalized-curve CV between 4.6\% and 5.5\%, comparable to random character noise (Table~\ref{tab:realistic_noise}). The non-universality thus holds across the character perturbations seen in real deployments, not just synthetic random substitution.

\paragraph{The split is robust to the choice of readout.} Reading the curve as $\text{JS}/\text{saturation}$ rather than $1-\text{JS}$ still leaves character noise above word noise (7.0\% versus 5.7\%), the ICC gives the same sixfold gap with no normalized-curve denominator at all, and word stays tighter at every individual noise level. The readout changes the absolute size of the character effect, never its sign.

All three measures agree, and agree with the embedding side: under word noise the LLM response is shared (CV 2.0\%, ICC 0.007), under character noise it carries model identity (CV 4.8\%, ICC 0.046). Decoder-only LLMs thus show the same split as sentence embeddings, despite a different architecture, readout, and objective, an agreement we trace in the next section to a cause both share: tokenization.

\section{Tokenization Causes the Split}
\label{sec:explanation}

\subsection{Character Noise Shatters the Token Sequence}
\label{sec:hierarchy}

Why does character noise scatter models when word noise does not? The mechanism sits upstream of either architecture: every model consumes a tokenized input, and the two noise types are not equally violent there. Dropping a word deletes its tokens and leaves the rest intact; substituting a single character forces the tokenizer to re-segment the surrounding word, often shattering one token into several unrelated subwords. Measured across five tokenizer families spanning both model sets (WordPiece and SentencePiece for the embeddings, BPE for the LLMs), a single character substitution induces \textbf{4--10$\times$} the token-edit distance of word dropout. At $\varepsilon{=}0.1$ the ratio is $6$--$10\times$ (Table~\ref{tab:hierarchy}; one character edit moves $2.7$--$3.7$ tokens, one word deletion $1.0$--$1.7$); it falls toward $4\times$ at higher rates as word dropout removes several words at once. The factor is largest for SentencePiece and smallest for byte-level BPE, but several-fold for every tokenizer tested.

This token-level amplification is what makes character noise look non-universal. The amount of shatter a character edit produces is a property of the tokenizer: different subword vocabularies re-segment a corrupted word differently, so models with different tokenizers diverge while models that share a tokenizer move together. They do: the character-noise CV across all nine embedding models is 14.2\%, but within the MiniLM-tokenizer group it falls to 5.6\%, and within the BERT-WordPiece group to 8.6\% (Figure~\ref{fig:tokenizer}). Holding the tokenizer fixed roughly halves the variation. Word dropout induces little shatter regardless of tokenizer, which is why word-noise response is shared even across models with different vocabularies. Re-tokenization shatter accounts for most of the asymmetry but not all of it: at a matched token-edit distance a word edit still displaces the representation about $3\times$ more than a character edit, so token count is a strong mediator, not a sufficient statistic.

\paragraph{A character-level backbone stays robust.} The tokenization mechanism makes a causal prediction: a backbone that tokenizes at the character level cannot shatter (one character maps to one token), so it should stay character-robust even under an identical objective. Contrastively training a character-level backbone (CANINE~\citep{clark2022canine}) alongside three subword backbones (BERT, RoBERTa, DistilBERT) with the same recipe confirms it: at $\varepsilon{=}0.1$ the character-level model retains 70\% of clean STS performance under character noise, versus 49--55\% (mean 53\%) for the subword models, while word-noise retention is comparable across all four (85--90\%).

\paragraph{Token perturbation is a continuous knob.}
\label{sec:granularity_axis}
If token perturbation is the controlling variable, then smoothly varying it, rather than the surface label of the noise, should smoothly move the cross-model spread. Two observations on the nine embedding models confirm this. First, across distinct noise types the spread tracks how much each perturbs the token sequence: word-scale perturbations, which barely shatter, are all universal (word dropout 2.2\%, word swap 1.0\%, synonym substitution 3.3\%), while every sub-word perturbation, which forces heavy re-tokenization, has a much higher inter-model CV (character substitution 14.2\%, byte dropout 17.7\%, character deletion 19.1\%, adjacent-bigram swap 19.4\%). Second, a single noise family with a tunable knob isolates the variable: corrupting contiguous blocks of $g$ characters at a fixed total rate, larger $g$ concentrates the corruption into fewer, longer runs and shatters less, so the induced token-edit distance per token falls monotonically from 0.73 at $g{=}1$ to 0.36 at $g{=}12$ and the inter-model CV falls in lockstep, from 14.3\% to 5.6\% (Figure~\ref{fig:gran_axis}).

\section{Reshaping the Noise Response by Training}
\label{sec:breaking}

If universality comes from a training objective constraining the noise response, we should be able to add a constraint by training and reshape the noise response. We show this in three settings: contrastive fine-tuning that moves a model onto the word-level universal curve, noise-augmented contrastive training for embeddings, and noise-augmented training for LLMs. Contrastive fine-tuning applies only to embeddings: its InfoNCE objective shapes a sentence-vector geometry, which decoder-only LLMs do not have.

\subsection{Embeddings: Fine-Tuning Onto the Word-Level Universal Curve}
\label{sec:phase}

\begin{figure}[t]
\centering
\includegraphics[width=0.74\columnwidth]{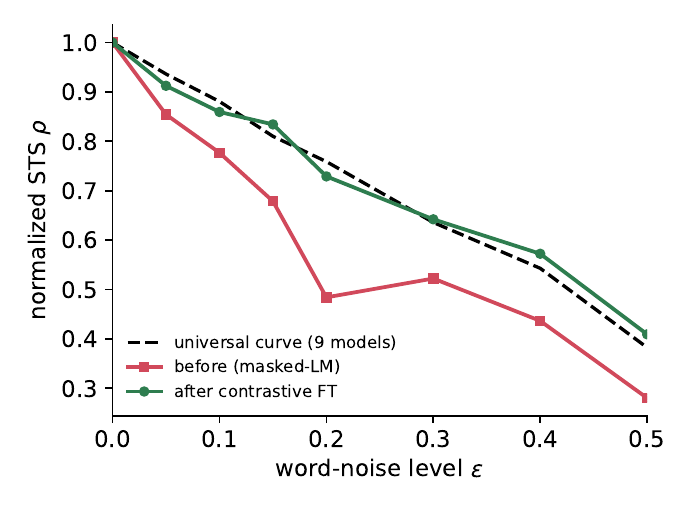}
\caption{A masked-LM BERT-base checkpoint (red) falls off the word-dropout universal curve (black dashed, mean of the nine sentence-trained models); after contrastive fine-tuning (green) it lies on it.}
\label{fig:phase}
\end{figure}

Contrastive fine-tuning alone can move a model onto the word-dropout universal curve. We verify this by fine-tuning a masked-LM BERT-base checkpoint with an InfoNCE objective (100K sentence pairs). As shown in Figure~\ref{fig:phase}, the model sits off the universal curve before fine-tuning and lies on it after. The move is fast: within about 200 steps (roughly one epoch) the normalized word-dropout curve reaches the universal shape and then stays fixed for the rest of training.

\subsection{Embeddings: Character Robustness by Augmentation}
\label{sec:shift_emb}

Character robustness can be added to an embedding model by training, even though its contrastive objective only rewards sentence-level invariance and never asks for it. We supply the missing invariance by augmentation: we train two BERT-base models with the same contrastive loss, differing only in that one replaces 50\% of its positive pairs with (clean, character-noised) versions. The augmented model degrades much more slowly under character noise (fitted decay rate $k$ from 8.4 to 6.2, retention at $\varepsilon{=}0.1$ from 52\% to 59\%), at a small cost to clean performance. The effect is scale-specific: character augmentation helps only character noise, and training on a mixture of character, word, and keyboard noise slows the character decay just as effectively while retaining robustness at the word scale (Figure~\ref{fig:curve_shift}).

\subsection{LLMs: Robustness by Augmentation}
\label{sec:shift_llm}

The same intervention applies to LLMs: we continue training a panel of models (unfreezing the top transformer blocks) on noisy text, then measure the JS response under both noise types. Noise-augmented training does raise robustness, mostly on the character scale. Either augmentation cuts the mean character-noise JS at $\varepsilon{=}0.1$ from 0.30 to 0.17 (lower JS means the noisy output stays closer to the clean one); word-noise JS, already low at 0.17 because word noise perturbs the token sequence little, barely moves. So the gain is real but concentrated where there is room to improve, the character scale.

\paragraph{Robustness rises, but universality does not.} Although augmentation reliably makes each model more robust, the augmented models do not collapse onto a tighter common curve: the cross-model character-noise CV actually rises, from 4.9\% at baseline to about 12\% after either augmentation. Augmentation acts on each model's already-divergent pretrained representation, improving robustness along a model-specific path rather than pulling the models together, the opposite of the shared contrastive objective. Universality is thus a property of that shared training-time constraint, not something post-hoc augmentation can install.

\paragraph{Practical payoff and its limit.} The robustness gain can transfer to real downstream tasks, but only where the task is sensitive to the full output distribution rather than the top-1 label. Noise-augmented SFT on Qwen2.5-1.5B (full) improves AG-News accuracy by $+$13\% at $\varepsilon{=}0.3$ with clean accuracy preserved (67\%$\to$68\%); LoRA SFT on Qwen2.5-7B-Instruct ($\sim$0.5\% of parameters) improves SST-2 by $+$5/$+$6/$+$19\% at $\varepsilon{=}0.1/0.2/0.3$ with clean accuracy unchanged (full per-noise-level results in Table~\ref{tab:sft_full}). Where a coarse top-1 label washes out the distribution shift, the accuracy gain is smaller, but the intervention still stabilizes the output distribution itself.

\section{Discussion}
\label{sec:discussion}

The split and its tokenization cause carry several practical consequences, which we draw out before turning to limitations.

\paragraph{Predicting robustness without noisy evaluation.}
In the universal word-noise regime every model follows the same normalized degradation curve, so a model's robustness can be read off that shared curve without evaluating it under noise at all. For embeddings the curve is the downstream metric itself (STS-B correlation, classification accuracy), so this directly answers how much task quality is lost at noise level $\varepsilon$. The prediction degrades gracefully outside the regime: even under the mildly non-universal LLM character noise, fitting the mean curve on all-but-one model predicts the held-out model's divergence with mean absolute error \textbf{3.5\%}. It is most useful where the output distribution matters, not just its argmax (calibration, sampling-based decoding, retrieval and logit fusion, distillation), and is not a promise about top-1 accuracy.

\paragraph{Matching the cleanup to the model.}
Input cleanup pays off most where a corruption induces large, tokenizer-dependent token perturbation, which for both architectures is character noise. The largest payoff is for embedding retrieval pipelines: they are robust to word edits but brittle to character corruption, so character-level normalization (spell-checking, OCR correction) is where the gains are. LLMs degrade more gently under character noise, so the same cleanup buys less; but because their character response is still tokenizer-dependent, a model that must survive a specific corruption channel cannot simply inherit another model's curve.

\paragraph{Universality lives at the output, not the internal geometry.} For LLMs, the tight word-noise response coexists with wide variation in middle-layer geometry: hidden-state isotropy differs substantially across families yet does not track generation robustness. This is consistent with the token-perturbation view, since the next-token objective constrains the output rather than the internal representation. The practical caveat is that reading an LLM's internal geometry need not tell you how its output degrades, even though representation geometry does track universality on the embedding side.

\paragraph{Limitations.}
Our study has three main boundaries. First, we perturb inputs only at the character and word scales; semantic perturbations such as paraphrase and content drift are untested. Second, the tokenization mechanism is an empirical regularity supported across regimes and interventions, not a first-principles derivation, and token-edit distance is a mediator rather than a sufficient statistic: at matched token distance a word edit still displaces a representation about three times more than a character edit, so factors beyond token count contribute. Third, our JS-divergence readout requires the full next-token distribution, which restricts it to open-weight models (proprietary API models expose only sampled tokens or truncated log-probabilities); we cover 18 open models up to 72B parameters, all in English. Extending the account to semantic perturbations, to models beyond 100B parameters, and to other languages is left for future exploration.


\section{Conclusion}
\label{sec:conclusion}
Whether neural language models degrade in the same way under noise depends on the scale of that noise. Under word-level noise, architecturally different sentence embeddings and 18 decoder-only LLMs trace nearly the same degradation curve; under character-level noise they separate. Two factors act together: a shared training objective is what makes different models converge to a common curve in the first place, and tokenization is what decides the scale at which that convergence holds, because a character edit forces tokenizer-specific re-segmentation that perturbs the token sequence far more than a dropped word. Neither factor is visible from a model's architecture or its internal geometry alone. Tokenization is thus a hidden variable behind the robustness and universality of language models.


\bibliographystyle{aaai2027}
\bibliography{references}

\begin{thebibliography}{29}
\providecommand{\natexlab}[1]{#1}

\bibitem[{Bahri et~al.(2020)Bahri, Kadmon, Pennington, Schoenholz,
  Sohl-Dickstein, and Ganguli}]{bahri2020statistical}
Bahri, Y.; Kadmon, J.; Pennington, J.; Schoenholz, S.~S.; Sohl-Dickstein, J.;
  and Ganguli, S. 2020.
\newblock Statistical mechanics of deep learning.
\newblock \emph{Annual review of condensed matter physics}, 11(1): 501--528.

\bibitem[{Belinkov and Bisk(2018)}]{belinkov2018synthetic}
Belinkov, Y.; and Bisk, Y. 2018.
\newblock Synthetic and Natural Noise Both Break Neural Machine Translation.
\newblock In \emph{International Conference on Learning Representations}.

\bibitem[{Bowman et~al.(2015)Bowman, Angeli, Potts, and
  Manning}]{bowman2015large}
Bowman, S.~R.; Angeli, G.; Potts, C.; and Manning, C.~D. 2015.
\newblock A large annotated corpus for learning natural language inference.
\newblock In \emph{Proceedings of the 2015 conference on empirical methods in
  natural language processing}, 632--642.

\bibitem[{Cer et~al.(2017)Cer, Diab, Agirre, Lopez-Gazpio, and
  Specia}]{cer2017semeval}
Cer, D.; Diab, M.; Agirre, E.; Lopez-Gazpio, I.; and Specia, L. 2017.
\newblock SemEval-2017 task 1: Semantic textual similarity multilingual and
  crosslingual focused evaluation.
\newblock In \emph{Proceedings of the 11th international workshop on semantic
  evaluation (SemEval-2017)}, 1--14.

\bibitem[{Clark et~al.(2022)Clark, Garrette, Turc, and
  Wieting}]{clark2022canine}
Clark, J.~H.; Garrette, D.; Turc, I.; and Wieting, J. 2022.
\newblock Canine: Pre-training an efficient tokenization-free encoder for
  language representation.
\newblock \emph{Transactions of the Association for Computational Linguistics},
  10: 73--91.

\bibitem[{Gao, Yao, and Chen(2021)}]{gao2021simcse}
Gao, T.; Yao, X.; and Chen, D. 2021.
\newblock Simcse: Simple contrastive learning of sentence embeddings.
\newblock In \emph{Proceedings of the 2021 conference on empirical methods in
  natural language processing}, 6894--6910.

\bibitem[{Huh et~al.(2024)Huh, Cheung, Wang, and Isola}]{huh2024platonic}
Huh, M.; Cheung, B.; Wang, T.; and Isola, P. 2024.
\newblock The platonic representation hypothesis.
\newblock \emph{arXiv preprint arXiv:2405.07987}.

\bibitem[{Jia and Liang(2017)}]{jia2017adversarial}
Jia, R.; and Liang, P. 2017.
\newblock Adversarial examples for evaluating reading comprehension systems.
\newblock In \emph{Proceedings of the 2017 conference on empirical methods in
  natural language processing}, 2021--2031.

\bibitem[{Kadanoff(1966)}]{kadanoff1966scaling}
Kadanoff, L.~P. 1966.
\newblock Scaling laws for Ising models near T c.
\newblock \emph{Physics Physique Fizika}, 2(6): 263.

\bibitem[{Kaplan et~al.(2020)Kaplan, McCandlish, Henighan, Brown, Chess, Child,
  Gray, Radford, Wu, and Amodei}]{kaplan2020scaling}
Kaplan, J.; McCandlish, S.; Henighan, T.; Brown, T.~B.; Chess, B.; Child, R.;
  Gray, S.; Radford, A.; Wu, J.; and Amodei, D. 2020.
\newblock Scaling laws for neural language models.
\newblock \emph{arXiv preprint arXiv:2001.08361}.

\bibitem[{Kornblith et~al.(2019)Kornblith, Norouzi, Lee, and
  Hinton}]{kornblith2019similarity}
Kornblith, S.; Norouzi, M.; Lee, H.; and Hinton, G. 2019.
\newblock Similarity of neural network representations revisited.
\newblock In \emph{International conference on machine learning}, 3519--3529.
  PMlR.

\bibitem[{Lu et~al.(2022)Lu, Bartolo, Moore, Riedel, and
  Stenetorp}]{lu2022fantastically}
Lu, Y.; Bartolo, M.; Moore, A.; Riedel, S.; and Stenetorp, P. 2022.
\newblock Fantastically ordered prompts and where to find them: Overcoming
  few-shot prompt order sensitivity.
\newblock In \emph{Proceedings of the 60th Annual Meeting of the Association
  for Computational Linguistics (Volume 1: Long Papers)}, 8086--8098.

\bibitem[{Merity et~al.(2016)Merity, Xiong, Bradbury, and
  Socher}]{merity2016pointer}
Merity, S.; Xiong, C.; Bradbury, J.; and Socher, R. 2016.
\newblock Pointer sentinel mixture models.
\newblock \emph{arXiv preprint arXiv:1609.07843}.

\bibitem[{Moradi and Samwald(2021)}]{moradi2021evaluating}
Moradi, M.; and Samwald, M. 2021.
\newblock Evaluating the robustness of neural language models to input
  perturbations.
\newblock In \emph{Proceedings of the 2021 Conference on Empirical Methods in
  Natural Language Processing}, 1558--1570.

\bibitem[{Morris et~al.(2020)Morris, Lifland, Yoo, Grigsby, Jin, and
  Qi}]{morris2020textattack}
Morris, J.; Lifland, E.; Yoo, J.~Y.; Grigsby, J.; Jin, D.; and Qi, Y. 2020.
\newblock Textattack: A framework for adversarial attacks, data augmentation,
  and adversarial training in nlp.
\newblock In \emph{Proceedings of the 2020 conference on empirical methods in
  natural language processing: System demonstrations}, 119--126.

\bibitem[{Pedregosa et~al.(2011)Pedregosa, Varoquaux, Gramfort, Michel,
  Thirion, Grisel, Blondel, Prettenhofer, Weiss, Dubourg
  et~al.}]{pedregosa2011scikit}
Pedregosa, F.; Varoquaux, G.; Gramfort, A.; Michel, V.; Thirion, B.; Grisel,
  O.; Blondel, M.; Prettenhofer, P.; Weiss, R.; Dubourg, V.; et~al. 2011.
\newblock Scikit-learn: Machine learning in Python.
\newblock \emph{the Journal of machine Learning research}, 12: 2825--2830.

\bibitem[{Pruthi, Dhingra, and Lipton(2019)}]{pruthi2019combating}
Pruthi, D.; Dhingra, B.; and Lipton, Z.~C. 2019.
\newblock Combating adversarial misspellings with robust word recognition.
\newblock In \emph{Proceedings of the 57th Annual Meeting of the Association
  for Computational Linguistics}, 5582--5591.

\bibitem[{Rajpurkar et~al.(2016)Rajpurkar, Zhang, Lopyrev, and
  Liang}]{rajpurkar2016squad}
Rajpurkar, P.; Zhang, J.; Lopyrev, K.; and Liang, P. 2016.
\newblock Squad: 100,000+ questions for machine comprehension of text.
\newblock In \emph{Proceedings of the 2016 conference on empirical methods in
  natural language processing}, 2383--2392.

\bibitem[{Reimers and Gurevych(2019)}]{reimers2019sentence}
Reimers, N.; and Gurevych, I. 2019.
\newblock Sentence-bert: Sentence embeddings using siamese bert-networks.
\newblock In \emph{Proceedings of the 2019 conference on empirical methods in
  natural language processing and the 9th international joint conference on
  natural language processing (EMNLP-IJCNLP)}, 3982--3992.

\bibitem[{Socher et~al.(2013)Socher, Perelygin, Wu, Chuang, Manning, Ng, and
  Potts}]{socher2013recursive}
Socher, R.; Perelygin, A.; Wu, J.; Chuang, J.; Manning, C.~D.; Ng, A.~Y.; and
  Potts, C. 2013.
\newblock Recursive deep models for semantic compositionality over a sentiment
  treebank.
\newblock In \emph{Proceedings of the 2013 conference on empirical methods in
  natural language processing}, 1631--1642.

\bibitem[{Thakur et~al.(2021)Thakur, Reimers, R{\"u}ckl{\'e}, Srivastava, and
  Gurevych}]{thakur2021beir}
Thakur, N.; Reimers, N.; R{\"u}ckl{\'e}, A.; Srivastava, A.; and Gurevych, I.
  2021.
\newblock Beir: A heterogenous benchmark for zero-shot evaluation of
  information retrieval models.
\newblock \emph{arXiv preprint arXiv:2104.08663}.

\bibitem[{Wang et~al.(2021)Wang, Xu, Wang, Gan, Cheng, Gao, Awadallah, and
  Li}]{wang2021adversarial}
Wang, B.; Xu, C.; Wang, S.; Gan, Z.; Cheng, Y.; Gao, J.; Awadallah, A.~H.; and
  Li, B. 2021.
\newblock Adversarial glue: A multi-task benchmark for robustness evaluation of
  language models.
\newblock \emph{arXiv preprint arXiv:2111.02840}.

\bibitem[{Wang et~al.(2022)Wang, Yang, Huang, Jiao, Yang, Jiang, Majumder, and
  Wei}]{wang2022text}
Wang, L.; Yang, N.; Huang, X.; Jiao, B.; Yang, L.; Jiang, D.; Majumder, R.; and
  Wei, F. 2022.
\newblock Text embeddings by weakly-supervised contrastive pre-training.
\newblock \emph{arXiv preprint arXiv:2212.03533}.

\bibitem[{Wang and Isola(2020)}]{wang2020understanding}
Wang, T.; and Isola, P. 2020.
\newblock Understanding contrastive representation learning through alignment
  and uniformity on the hypersphere.
\newblock In \emph{International conference on machine learning}, 9929--9939.
  PMLR.

\bibitem[{Wei et~al.(2022)Wei, Tay, Bommasani, Raffel, Zoph, Borgeaud,
  Yogatama, Bosma, Zhou, Metzler et~al.}]{wei2022emergent}
Wei, J.; Tay, Y.; Bommasani, R.; Raffel, C.; Zoph, B.; Borgeaud, S.; Yogatama,
  D.; Bosma, M.; Zhou, D.; Metzler, D.; et~al. 2022.
\newblock Emergent abilities of large language models.
\newblock \emph{arXiv preprint arXiv:2206.07682}.

\bibitem[{Wilson(1971)}]{wilson1971renormalization}
Wilson, K.~G. 1971.
\newblock Renormalization group and critical phenomena. I. Renormalization
  group and the Kadanoff scaling picture.
\newblock \emph{Physical review B}, 4(9): 3174.

\bibitem[{Xiao et~al.(2024)Xiao, Liu, Zhang, Muennighoff, Lian, and
  Nie}]{xiao2024c}
Xiao, S.; Liu, Z.; Zhang, P.; Muennighoff, N.; Lian, D.; and Nie, J.-Y. 2024.
\newblock C-pack: Packed resources for general chinese embeddings.
\newblock In \emph{Proceedings of the 47th international ACM SIGIR conference
  on research and development in information retrieval}, 641--649.

\bibitem[{Zhang, Zhao, and LeCun(2015)}]{zhang2015character}
Zhang, X.; Zhao, J.; and LeCun, Y. 2015.
\newblock Character-level convolutional networks for text classification.
\newblock \emph{Advances in neural information processing systems}, 28.

\bibitem[{Zou et~al.(2023)Zou, Wang, Carlini, Nasr, Kolter, and
  Fredrikson}]{zou2023universal}
Zou, A.; Wang, Z.; Carlini, N.; Nasr, M.; Kolter, J.~Z.; and Fredrikson, M.
  2023.
\newblock Universal and transferable adversarial attacks on aligned language
  models.
\newblock \emph{arXiv preprint arXiv:2307.15043}.

\end{thebibliography}

\appendix
\setcounter{secnumdepth}{1}

\section{Normalization Does Not Manufacture the Agreement}
\label{app:normalization}

Here ``universal'' refers to the shape of the normalized degradation curve $\rho(\varepsilon)/\rho(0)$, not to the absolute performance drop: two models can lose different absolute amounts of quality yet follow the same decay law once each is referred to its own clean baseline. A natural worry is that normalization could manufacture agreement, taking, say, one model that drops 40\% and another that drops 20\% and rescaling them into coincidence. It does not, for two reasons. First, the rescaling is mild: the clean baselines themselves span a narrow range ($\rho(0)$ from 0.80 to 0.88, CV 2.8\%), so dividing by $\rho(0)$ moves the curves little. Second, and decisively, the word-noise agreement is already present before any normalization. Table~\ref{tab:raw_retention} reports the raw STS-B retention (no division by $\rho(0)$) across the nine models: under word dropout the models span barely two percentage points at each noise level (87--89\% at $\varepsilon{=}0.1$, 63--65\% at $\varepsilon{=}0.3$), whereas under character noise they fan out across roughly twelve points (51--63\% at $\varepsilon{=}0.1$). The ``40\% versus 20\%'' scenario simply does not occur under word noise; it does under character noise, which is why we report that regime as non-universal rather than normalizing the disagreement away.

This concern applies only to the CV. The two other measures do not divide by $\rho(0)$: the decay exponent $k$ is a relative rate fitted to each curve, and the ICC is computed from per-pair cosine retention, which already starts at one. Neither can be inflated or deflated by the choice of clean-value normalization, so the word/character split they report is immune to this worry by construction.

\begin{table}[t]
\small
\centering
\caption{The word-noise agreement is present before normalization does any work. For each noise level we list the range, across the nine models, of STS-B quality retained relative to clean ($\rho(\varepsilon)/\rho(0)$); since the clean baselines $\rho(0)$ themselves span only $0.80$--$0.88$, the absolute scores cluster just as tightly. Under word dropout the models stay within a two-point band at every level; under character noise they already fan out roughly six times more.}
\label{tab:raw_retention}
\begin{tabular}{lcc}
\toprule
Noise & Across-model retention range & Spread \\
\midrule
Word, $\varepsilon{=}0.1$ & 87--89\% & 2 pts \\
Word, $\varepsilon{=}0.3$ & 63--65\% & 2 pts \\
Character, $\varepsilon{=}0.1$ & 51--63\% & 12 pts \\
\bottomrule
\end{tabular}
\end{table} Where a choice of rescaling could turn a large spread into a small one, we do not use it to adjudicate universality, and rely instead on the self-normalizing downstream metric (Spearman $\rho$, bounded in $[-1,1]$), which requires no such choice.

\section{The Within-Model Reference Must Be Per-Input, Not Dataset-Level}
\label{app:ruler}

The ICC asks what fraction of the noise response is explained by model identity rather than by the input and the noise draw, so it needs a within-model variance: fix a model, fix an input, redraw only the noise seed, and see how much the response moves. How this within-model term is measured decides whether the ICC means anything, and the natural-looking choice is wrong.

The wrong choice is a dataset-level within-model term: take the model's aggregate curve (for embeddings, one Spearman $\rho$ over the whole evaluation set at each $\varepsilon$), redraw the noise, and record how much that aggregate wobbles. This wobble is not an intrinsic property of the model; it is the estimation error of the aggregate, and it shrinks like $1/\sqrt{N}$ as the evaluation set grows (about 7\% at $N{=}50$ pairs, 2.3\% at the 1{,}379 pairs of STS-B, $\sim$0.7\% at ten times that). An ICC built on it would inherit that $1/\sqrt{N}$ dependence and could be driven to any value by the choice of evaluation size, so ``how distinguishable are these models'' would be answered by how large a benchmark one happened to use. It is the same mistake as judging whether two people are the same height from whether their difference falls within one tape measure's error bars: a finer tape flips the answer.

The correct within-model term is measured at the level of a single input: fix one sentence (for LLMs) or one sentence pair (for embeddings), redraw the noise, and take the variance of that one item's response. This is an intrinsic property of the item and does not shrink with $N$. It requires a per-input readout, which the LLM side has natively (per-sentence $1-\text{JS}$) and which we construct for embeddings as the per-pair cosine retention $\cos(\text{clean}, \text{noisy})$ rather than the dataset-level $\rho$. With this term the ICC is sample-size-independent and directly comparable across the two architectures, and it is what we report: word noise has a much smaller ICC than character noise for both embeddings and LLMs.

For completeness, the dataset-level $\rho$ readout used for the main embedding CV does not admit a per-input within-model term (a single pair has no $\rho$), which is why the embedding ICC uses the per-pair cosine readout instead. The two embedding readouts agree on the ordering and on the several-fold word/character gap, so the choice of readout does not affect the conclusion.

\section{Shared-Exponent Test and the Non-Universal Amplitude}
\label{app:exponent}

This appendix details the universality-class criterion from the Setup: a set of models is one class when the decay exponents $k_m$ of $c_m(\varepsilon)=A_m e^{-k_m\varepsilon}+(1-A_m)$ coincide within uncertainty. The exponent is read cleanly only when the amplitude $A_m$ is common across models; otherwise the fit trades $A$ against $k$ and the raw $k_m$ inherits the spread of $A_m$. For the embedding models $A_m\approx1.00$ (the embedding results), so the raw $k_m$ is unambiguous and we report it directly. For the LLMs $A_m$ ranges over $0.53$--$0.75$ and is model-specific, so we first remove it by saturation normalization (divide each curve's noise-sensitive part by its own plateau height) and compare the resulting exponent.

\paragraph{The amplitude is non-universal.} The amplitude $A_m$ is a per-model prefactor, the role a non-universal amplitude plays in a physical universality class; it is not the universal quantity, so its variation across models does not bear on the universality-class verdict, which rests on the exponent. Under word noise $A_m$ does carry a scale dependence (larger models have a smaller $A_m$, Spearman $\rho{=}{-}0.66$, $p{=}0.01$), while under character noise it is uncorrelated with size ($\rho{=}0.24$, $p{=}0.33$). Either way, it is the exponent, not the amplitude, that is shared, which is why we divide the amplitude out before comparing exponents.

\paragraph{Joint-bootstrap exponent dispersion.} Table~\ref{tab:exponent} reports the across-model coefficient of variation of the (amplitude-controlled) exponent, with a joint bootstrap that refits $A_m$ and $k_m$ together on each resample so that the amplitude's estimation error propagates into the exponent's confidence interval. Word exponents are several times tighter than character exponents in both architectures, and the word and character intervals do not overlap, matching the CV and ICC verdicts.

\begin{table}[t]
\small
\centering
\caption{Across-model dispersion of the decay exponent (CV of $k_m$, \%), with bootstrap 95\% CIs. Embedding uses the raw exponent (with a bootstrap-over-models CI); LLM uses the amplitude-controlled (saturation-normalized) exponent (joint-bootstrap CI, refitting $A_m$ and $k_m$ on each resample). In every cell the word exponent is far more shared than the character exponent, and the word/character CIs are disjoint.}
\label{tab:exponent}
\begin{tabular}{llcc}
\toprule
Architecture & Scale & Exponent CV & 95\% CI \\
\midrule
Embedding & word & \textbf{4.3\%} & [2.2, 5.0] \\
Embedding & char & 13.1\% & [5.3, 17.7] \\
\midrule
LLM & word & \textbf{5.3\%} & [4.8, 6.4] \\
LLM & char & 15.2\% & [12.6, 20.4] \\
\bottomrule
\end{tabular}
\end{table}

\section{Detailed Embedding Model Specifications}
\label{app:models}

Table~\ref{tab:all_models_app} lists the architecture, pooling, embedding dimension, training objective, and the fitted character decay rate $k_\text{char}$ for the 13 embedding models. The nine sentence-trained models span four backbone architectures (MiniLM, MPNet, BERT, T5 encoder), three embedding dimensions (384, 768, 1024), and both mean and CLS pooling, yet all join the common word-dropout curve, while the four non-sentence-trained baselines (masked-LM or decoder) do not. The 18 decoder-only LLMs are listed in Table~\ref{tab:llm_models}, with per-model character-noise JS in Table~\ref{tab:llm_univ}.

\begin{table*}[t]
\small
\centering
\caption{Embedding model inventory. The nine sentence-trained models differ in backbone, dimension, and pooling but join the common word-dropout curve (``word-universal''); the four baselines do not. $k_\text{char}$ is the fitted character-noise decay rate on STS-B and varies widely even among the word-universal models, reflecting the tokenizer dependence of character-level response.}
\label{tab:all_models_app}
\begin{tabular}{llcccccc}
\toprule
Model & Backbone & Params & Dim & Pooling & Training & $k_\text{char}$ & Word-universal \\
\midrule
all-MiniLM-L6-v2 & MiniLM & 23M & 384 & mean & contrastive & 8.38 & yes \\
all-MiniLM-L12-v2 & MiniLM & 33M & 384 & mean & contrastive & 8.34 & yes \\
all-mpnet-base-v2 & MPNet & 110M & 768 & mean & contrastive & 6.54 & yes \\
bge-base-en-v1.5 & BERT & 110M & 768 & CLS & contrastive & 8.70 & yes \\
bge-large-en-v1.5 & BERT & 335M & 1024 & CLS & contrastive & 7.97 & yes \\
e5-base-v2 & BERT & 110M & 768 & mean & contrastive & 7.41 & yes \\
e5-large-v2 & BERT & 335M & 1024 & mean & contrastive & 7.21 & yes \\
gtr-t5-base & T5 encoder & 110M & 768 & mean & contrastive & 5.68 & yes \\
nomic-embed-text-v1.5 & BERT (RoPE) & 137M & 768 & mean & contrastive & 8.28 & yes \\
\midrule
bert-base-uncased & BERT & 110M & 768 & CLS & masked LM & 12.91 & no \\
bert-large-uncased & BERT & 335M & 1024 & CLS & masked LM & 24.21 & no \\
roberta-base & RoBERTa & 125M & 768 & CLS & masked LM & 28.82 & no \\
e5-mistral-7b & Mistral & 7B & 4096 & last-token & contrastive (dec.) & 10.01 & no \\
\bottomrule
\end{tabular}
\end{table*}

\begin{table}[t]
\small
\centering
\caption{The 18 decoder-only LLMs, spanning nine families, two Qwen generations, base/instruct/chat variants, and a 48$\times$ range in scale (1.5B--72B). All use byte-pair-encoding (BPE) tokenizers. Per-model character-noise JS is in Table~\ref{tab:llm_univ}.}
\label{tab:llm_models}
\begin{tabular}{llc}
\toprule
Model & Family & Params \\
\midrule
GPT2-XL & GPT-2 & 1.5B \\
Qwen2.5-1.5B & Qwen2.5 & 1.5B \\
Qwen2.5-3B-Instruct & Qwen2.5 & 3B \\
Qwen2.5-7B & Qwen2.5 & 7B \\
Qwen2.5-7B-Instruct & Qwen2.5 & 7B \\
Qwen2.5-14B-Instruct & Qwen2.5 & 14B \\
Qwen2.5-32B-Instruct & Qwen2.5 & 32B \\
Qwen2.5-72B-Instruct & Qwen2.5 & 72B \\
Qwen3-1.7B & Qwen3 & 1.7B \\
Qwen3-8B & Qwen3 & 8B \\
Phi-2 & Phi & 2.7B \\
Phi-3-mini & Phi & 3.8B \\
Llama-2-7B & Llama-2 & 7B \\
Llama-2-7B-Chat & Llama-2 & 7B \\
Llama-3-8B-Instruct & Llama-3 & 8B \\
Mistral-7B-Instruct & Mistral & 7B \\
DeepSeek-7B-Chat & DeepSeek & 7B \\
Yi-1.5-9B-Chat & Yi & 9B \\
\bottomrule
\end{tabular}
\end{table}

\begin{table}[t]
\small
\centering
\caption{Raw JS divergence between clean and noisy next-token distributions at $\varepsilon = 0.1$ under character noise, for the 18 LLMs. Both the absolute values and the normalized curve shapes spread across models (normalized-curve CV 4.8\%, versus 2.0\% under word noise).}
\label{tab:llm_univ}
\begin{tabular}{lcc}
\toprule
Model & Params & JS@$\varepsilon$=0.1 \\
\midrule
Llama-2-7B & 7B & 0.241 \\
Mistral-7B-Instruct & 7B & 0.249 \\
Qwen3-1.7B & 1.7B & 0.253 \\
Llama-2-7B-Chat & 7B & 0.262 \\
Phi-3-mini & 3.8B & 0.265 \\
Qwen2.5-7B & 7B & 0.266 \\
DeepSeek-7B-Chat & 7B & 0.268 \\
Qwen3-8B & 8B & 0.281 \\
Qwen2.5-1.5B & 1.5B & 0.285 \\
Llama-3-8B-Instruct & 8B & 0.293 \\
Phi-2 & 2.7B & 0.298 \\
Qwen2.5-3B-Instruct & 3B & 0.306 \\
Yi-1.5-9B-Chat & 9B & 0.315 \\
Qwen2.5-7B-Instruct & 7B & 0.318 \\
GPT2-XL & 1.5B & 0.334 \\
Qwen2.5-14B-Instruct & 14B & 0.350 \\
Qwen2.5-72B-Instruct & 72B & 0.362 \\
Qwen2.5-32B-Instruct & 32B & 0.363 \\
\bottomrule
\end{tabular}
\end{table}

\section{Cross-Architecture Experiment: Per-Model Numbers}
\label{app:arch}

Table~\ref{tab:architecture} gives the per-model word-noise retention behind the before/after collapse in the objective-not-architecture experiment in the main paper. All eight encoders are contrastively trained from their pretrained checkpoints under one identical recipe (InfoNCE, $\tau{=}0.05$, 50K SNLI entailment pairs, 5K steps). Before training their word-dropout curves are scattered (inter-model CV 14.3\%, averaged over ten noise seeds per level to remove single-draw jitter); after training they collapse onto a common curve (2.8\%), while character noise tightens only partly (29.7\% to 13.0\%) and stays non-universal.

\begin{table}[t]
\small
\centering
\caption{Per-model results for the cross-architecture experiment. The bottom rows give the before/after inter-model CV; the per-model column is word retention (fraction of clean STS performance kept at $\varepsilon{=}0.2$ under word dropout) after training.}
\label{tab:architecture}
\begin{tabular}{llc}
\toprule
Architecture & Pretraining paradigm & Word ret.\ @\,0.2 \\
\midrule
BERT-base & masked LM & 0.73 \\
RoBERTa-base & masked LM (byte-BPE) & 0.75 \\
DistilBERT & masked LM (distilled) & 0.74 \\
ALBERT & param.\ sharing & 0.75 \\
ELECTRA & replaced-token det.\ & 0.72 \\
DeBERTa & disentangled attn.\ & 0.76 \\
XLNet & permutation LM & 0.77 \\
T5 encoder & enc--dec span & 0.74 \\
\midrule
\multicolumn{2}{l}{\textbf{Word CV (before $\to$ after)}} & \textbf{14.3\% $\to$ 2.8\%} \\
\multicolumn{2}{l}{\textbf{Char CV (before $\to$ after)}} & \textbf{29.7\% $\to$ 13.0\%} \\
\bottomrule
\end{tabular}
\end{table}

\section{Computing Infrastructure}
\label{app:compute}

All experiments were run on NVIDIA A100-SXM4 GPUs (40\,GB each) under Linux, with CUDA 12.4. Every measurement in the paper fits on a single GPU; the largest models (32B and 72B) use one A100 with half-precision weights, and no experiment requires multi-GPU or model parallelism. The software stack is Python 3.10 with PyTorch 2.6.0, Transformers 4.44.2, sentence-transformers 5.3.0, and Datasets 2.19.0 for models and data, and NumPy 2.2.6, SciPy 1.15.3, scikit-learn 1.4.2, and Matplotlib 3.9.0 for the analysis and figures. The noise perturbations, curve fitting, bootstrap and jackknife procedures, and all plotting use only these standard libraries. All models and datasets are the public releases cited in the main text, loaded through the Hugging Face Hub.

\section{Embedding Degradation Curves for All Noise Types}
\label{app:emb_all_noise}

\begin{figure*}[t]
\centering
\includegraphics[width=\textwidth]{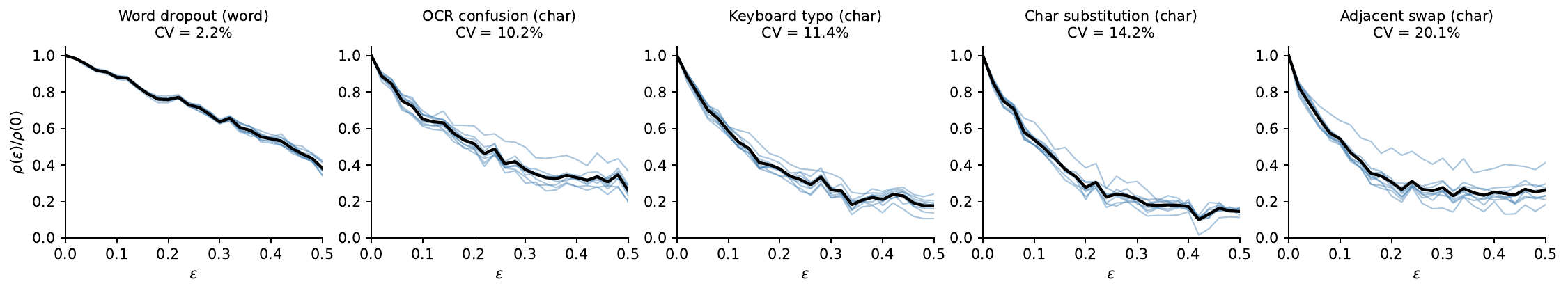}
\caption{Normalized STS-B degradation curves for the nine sentence-trained embedding models under five noise types (each thin line is one model; black is the mean). Curves collapse under word dropout (CV 2.2\%) and diverge under all character-level noise (CV 10--20\%).}
\label{fig:emb_all_noise}
\end{figure*}

Figure~\ref{fig:emb_all_noise} shows the full normalized degradation curves of the nine sentence-trained models under all five noise types, the visual counterpart to the CV values in Table~\ref{tab:emb_noise_cv}. The nine curves visibly coincide only under word dropout; under every character-level noise type (OCR, keyboard, substitution, swap) they fan out, with the spread growing as the perturbation becomes finer-grained.

\section{Stretched-Exponential Analysis}
\label{app:stretched}

The functional form $f(\varepsilon) = (1-c)\exp(-(k\varepsilon)^\beta) + c$ with free stretching exponent $\beta$ provides a stringent test of the memoryless (independent-hit) hypothesis.

Under character noise: $\beta = 0.958$, 95\% CI $[0.899, 1.017]$. The $\beta = 1$ hypothesis cannot be rejected ($p = 0.16$), confirming that character corruption acts as independent, memoryless hits, consistent with a Poisson process where the encoder processes character-level noise token-by-token.

Under word noise: $\beta = 1.181$, 95\% CI $[1.028, 1.334]$. The $\beta = 1$ hypothesis is decisively rejected ($p < 10^{-4}$), indicating that word removal is cooperative: the first few dropped words do disproportionate damage before saturating. This reflects that word dropout directly attacks the compositional scaffold of the sentence, where early losses cascade.

\section{Why Post-Hoc Intervention Fails}
\label{app:noise_aug}

This appendix supports the claim that noise robustness must be modified during training rather than after it. We attempted to shift the noise response of an already-trained model by post-hoc fine-tuning:

\begin{itemize}
\item Naive post-hoc fine-tuning (lr $=10^{-5}$, last 4 layers, character-noise pairs) shows an apparent improvement in the noise metric but causes catastrophic forgetting: downstream task accuracy collapses to near 0\%.
\item With a smaller lr $=10^{-6}$, clean accuracy is preserved but the robustness gain is marginal ($+$5--8 points on SST-2).
\item The gain depends on baseline robustness: already-robust models (top-1@0.1 $>$ 60\%) show no improvement, while weaker models (top-1@0.1 $<$ 50\%) gain $+$7--16 points on top-1 match.
\end{itemize}

These results motivate the training-time interventions in the main text: for embeddings, modifying the contrastive training data; for LLMs, augmenting noise during instruction tuning rather than after pre-training.

\section{Noise-Augmented Instruction Tuning: Full Results}
\label{app:sft}

This appendix gives the per-noise-level task accuracies behind the practical-payoff claims in the main text. The intervention is noise-augmented supervised fine-tuning (SFT): during instruction tuning, a fraction of training inputs is corrupted with character noise so the model learns to produce the clean-input answer from a noised prompt. We run it in two regimes: full-parameter SFT on Qwen2.5-1.5B, and LoRA SFT on Qwen2.5-7B-Instruct (rank-16 adapters, $\sim$0.5\% of parameters). Both are evaluated by top-1 task accuracy on AG-News (4-way) and SST-2 (binary) as the input is corrupted at increasing character-noise rate $\varepsilon$, against a standard-SFT baseline trained identically but on clean inputs.

\begin{table}[t]
\small
\centering
\caption{Top-1 accuracy under character noise, standard SFT vs.\ noise-augmented SFT, at each noise level. Clean accuracy ($\varepsilon{=}0$) is preserved while accuracy under noise improves, the gain growing with $\varepsilon$. Full-parameter SFT on Qwen2.5-1.5B (top) and LoRA SFT on Qwen2.5-7B-Instruct (bottom).}
\label{tab:sft_full}
\setlength{\tabcolsep}{4pt}
\begin{tabular}{llcccccc}
\toprule
Model / task & SFT & \multicolumn{6}{c}{accuracy at $\varepsilon=$} \\
 & & 0.0 & 0.05 & 0.1 & 0.15 & 0.2 & 0.3 \\
\midrule
\multicolumn{8}{l}{Qwen2.5-1.5B, full SFT} \\
AG-News & standard & 0.67 & 0.68 & 0.67 & 0.63 & 0.62 & 0.46 \\
AG-News & noise-aug & 0.68 & 0.65 & 0.69 & 0.69 & 0.67 & \textbf{0.59} \\
SST-2 & standard & 0.90 & 0.85 & 0.79 & 0.81 & 0.71 & 0.57 \\
SST-2 & noise-aug & 0.87 & 0.83 & 0.77 & 0.76 & 0.68 & \textbf{0.62} \\
\midrule
\multicolumn{8}{l}{Qwen2.5-7B-Instruct, LoRA SFT} \\
AG-News & standard & 0.76 & 0.78 & 0.76 & 0.78 & 0.76 & 0.61 \\
AG-News & noise-aug & 0.72 & 0.73 & 0.73 & 0.72 & 0.71 & \textbf{0.66} \\
SST-2 & standard & 0.91 & 0.88 & 0.83 & 0.77 & 0.71 & 0.55 \\
SST-2 & noise-aug & 0.91 & 0.89 & \textbf{0.88} & 0.79 & \textbf{0.77} & \textbf{0.74} \\
\bottomrule
\end{tabular}
\end{table}

Table~\ref{tab:sft_full} shows the pattern quoted in the main text: clean accuracy is essentially unchanged ($\varepsilon{=}0$ within a point of the baseline), while accuracy under noise improves, and the improvement widens as $\varepsilon$ grows, reaching $+$13 points on AG-News and $+$19 points on SST-2 at $\varepsilon{=}0.3$. The boundary noted in the main text is also visible here: on AG-News the gain is modest at small $\varepsilon$ (a top-1 label is insensitive to the fine-grained distribution shift that the JS metric captures), and emerges clearly only once the noise is heavy enough to flip labels. The effect is consistent across both a full-parameter small model and a parameter-efficient adapter on a larger one, so the robustness transfer is not specific to model size or to how much of the network is updated.

\section{Cross-Dataset Validation}
\label{app:cross_dataset}

To check that the word-level universality of embeddings is not specific to the STS-B similarity task, we evaluate the nine sentence-trained models under word dropout on two further tasks: classification on AG News (nearest-centroid accuracy) and clustering on 20 Newsgroups (silhouette score). Applying the same normalized-curve CV recipe, the word-dropout response is universal on both: inter-model CV is 1.7\% on AG News and 7.6\% on 20 Newsgroups, alongside the 2.2\% on STS-B. (CV must be computed on per-curve-normalized degradation shapes; comparing raw silhouette values, which differ several-fold in absolute scale across models and sit near zero, spuriously inflates the apparent spread.) Word-level universality is thus a property of the contrastive representation, not of any single evaluation task.

\paragraph{A retrieval benchmark.} Because these embeddings are most often deployed for retrieval, we also evaluate on a standard retrieval benchmark, SciFact from BEIR~\citep{thakur2021beir}, corrupting the query (the realistic case of a clean index and a noisy user query) and measuring nDCG@10 degradation over the nine models. The pattern matches STS-B, and is if anything sharper: the word-dropout response is universal (inter-model CV \textbf{2.2\%}, bootstrap 95\% CI $[1.3, 2.6]$), while the character-noise response spreads the models apart (CV \textbf{17.1\%}, CI $[9.9, 21.1]$), the two intervals well separated. Under character noise retrieval quality collapses (nDCG@10 retention of only 0.56--0.77 at $\varepsilon{=}0.1$, versus 0.92--0.97 under word dropout), consistent with the tokenization mechanism acting the same way whether the readout is similarity, classification, clustering, or retrieval.

For LLMs, the character-noise response (normalized-curve CV 4.8\%) is measured over a 1000-sentence set pooled from five domains (see the Setup section), so the cross-model spread already reflects cross-domain behavior and is not an artifact of a single domain.

\paragraph{The LLM word/character split is robust to the normalization choice.} The main text reports LLM curves as retained agreement $1-\text{JS}$, normalized by its clean value, so that the curve decays from one like the embedding score. The alternative is to keep the raw $\text{JS}$ divergence and normalize by each model's saturation level. The split is the same either way: word noise tighter than character noise. Under $1-\text{JS}$ the cross-model CV is 2.0\% (word) and 4.8\% (character); under $\text{JS}/\text{saturation}$ it is 5.7\% (word) and 7.0\% (character). In both cases the character CV exceeds the word CV, so the conclusion that word noise is the universal regime for LLMs while character noise is the non-universal one does not depend on which normalization is used. We report $1-\text{JS}$ in the main text only because it puts both regimes on a single decaying form.

\section{Additional Figures and Tables}
\label{app:supp}

\begin{figure}[htbp]
\centering
\includegraphics[width=0.78\columnwidth]{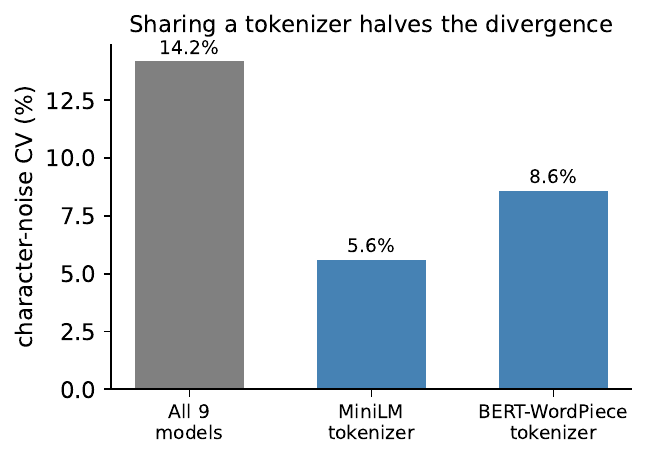}
\caption{The tokenizer drives character-level divergence. The character-noise CV across embedding models drops from 14.2\% (all nine) to 5.6--8.6\% within a shared-tokenizer group.}
\label{fig:tokenizer}
\end{figure}

\begin{figure}[htbp]
\centering
\includegraphics[width=0.82\columnwidth]{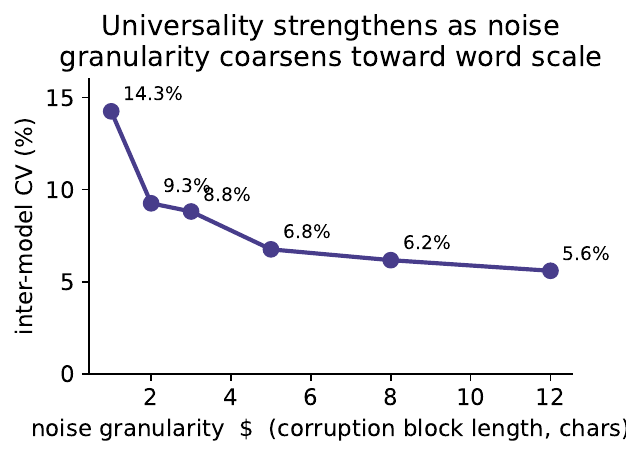}
\caption{Cross-model spread tracks token perturbation. A single block-corruption family with tunable block length $g$ (characters) sweeps the induced token-edit distance: as $g$ grows, the same corruption rate touches fewer words and shatters less (token-edit distance per token falls from 0.73 at $g{=}1$ to 0.36 at $g{=}12$), and the inter-model CV falls monotonically in lockstep, from 14.3\% at $g{=}1$ to 5.6\% at $g{=}12$.}
\label{fig:gran_axis}
\end{figure}

\begin{figure}[htbp]
\centering
\includegraphics[width=0.82\columnwidth]{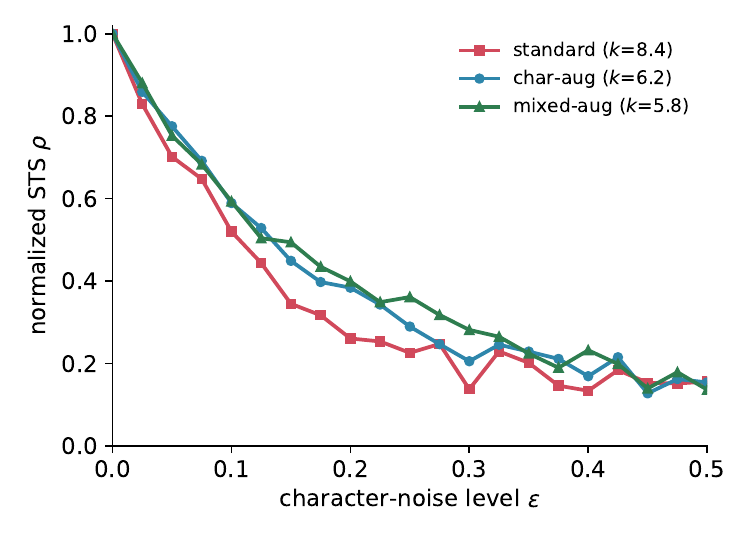}
\caption{Noise-augmented contrastive training slows the character-noise decay. Under character noise, the standard model ($k{=}8.4$) degrades fastest; char- and mixed-augmented training ($k{=}6.2$, $5.8$) decay more slowly. Curves are normalized STS $\rho$ across the embedding models.}
\label{fig:curve_shift}
\end{figure}

This appendix collects figures and tables referenced from the main text. Table~\ref{tab:band_verdict} summarizes the word/character split across both architectures. Table~\ref{tab:emb_noise_cv} gives the embedding cross-model CV under all five noise types. Table~\ref{tab:hierarchy} gives the per-tokenizer token-edit distances behind the 4--10$\times$ amplification. Table~\ref{tab:realistic_noise} lists the cross-model CV under each structured character-noise type for LLMs. Table~\ref{tab:llm_aug} reports the effect of noise-augmented training on the LLM JS response. Figure~\ref{fig:tokenizer} shows that holding the tokenizer fixed roughly halves the embedding character-noise CV. Figure~\ref{fig:gran_axis} shows the single block-corruption knob that sweeps the cross-model spread by tuning token perturbation. Figure~\ref{fig:curve_shift} shows noise-augmented contrastive training slowing the embedding character-noise decay.

\begin{table}[htbp]
\small
\centering
\caption{The word/character split across both architectures. For each architecture and noise scale we report the cross-model CV of the normalized degradation curve. Word noise collapses to near 2\% for both, two model classes that share nothing but the training signal; character noise is far larger for embeddings and milder for LLMs. The last row gives, for comparison, the same architecturally diverse encoders measured before the shared objective: 14.3\%, far more scattered.}
\label{tab:band_verdict}
\setlength{\tabcolsep}{4pt}
\begin{tabular}{llcc}
\toprule
Architecture & Scale & Cross-model CV & Reading \\
\midrule
Embedding & word & 2.2\% & universal \\
Embedding & char & 14.2\% & \textbf{non-universal} \\
\midrule
LLM & word & 2.0\% & universal \\
LLM & char & 4.8\% & \textbf{partly broken} \\
\midrule
\multicolumn{2}{l}{untrained encoders} & 14.3\% & (before obj.) \\
\bottomrule
\end{tabular}
\end{table}

\begin{table}[htbp]
\small
\centering
\caption{Word-level noise response is universal across the nine sentence-trained models; character-level noise spreads them apart. Pointwise CV of the normalized STS-B degradation curve (averaged over $\varepsilon>0$).}
\label{tab:emb_noise_cv}
\begin{tabular}{lcc}
\toprule
Noise type & Scale & CV \\
\midrule
Word dropout & word & \textbf{2.2\%} \\
\midrule
Character substitution & character & 14.2\% \\
Keyboard typo (QWERTY) & character & 11.4\% \\
OCR confusion & character & 10.2\% \\
Adjacent swap & character & 20.1\% \\
\bottomrule
\end{tabular}
\end{table}

\begin{table}[htbp]
\small
\centering
\caption{Noise-augmented training in LLMs (mean JS@$\varepsilon{=}0.1$, $\downarrow$; lower is more robust) raises robustness mostly on the character scale: either augmentation cuts character-noise JS from 0.30 to 0.17, while word-noise JS is already low and barely moves. Mean over the augmented panel (9 models for character augmentation, 6 for word).}
\label{tab:llm_aug}
\begin{tabular}{lcc}
\toprule
 & \multicolumn{2}{c}{robustness (JS@0.1, $\downarrow$)} \\
Augmentation & char & word \\
\midrule
None (baseline) & 0.30 & 0.17 \\
Character-aug & \textbf{0.17} & 0.15 \\
Word-aug & \textbf{0.17} & 0.15 \\
\bottomrule
\end{tabular}
\end{table}

\begin{table}[htbp]
\small
\centering
\caption{Normalized token-edit distance (Levenshtein on token-id sequences $\div$ clean length) at $\varepsilon{=}0.1$ and the character/word ratio, for five tokenizer families spanning both model sets (WordPiece and SentencePiece for the embedding models, BPE for the LLMs). A character edit perturbs the token sequence several-fold more than a word edit for every tokenizer.}
\label{tab:hierarchy}
\setlength{\tabcolsep}{4.5pt}
\begin{tabular}{lccc}
\toprule
Tokenizer & Word ED & Char ED & Char/Word \\
\midrule
\multicolumn{4}{l}{Embedding models} \\
WordPiece (MiniLM, BGE) & 0.09 & 0.72 & 7.8$\times$ \\
SentencePiece (GTR-T5)  & 0.09 & 0.86 & 9.8$\times$ \\
\midrule
\multicolumn{4}{l}{LLMs} \\
BPE (GPT-2)             & 0.11 & 0.71 & 6.8$\times$ \\
BPE (Qwen2.5)           & 0.11 & 0.66 & 6.2$\times$ \\
BPE (Llama)             & 0.11 & 0.69 & 6.6$\times$ \\
\bottomrule
\end{tabular}
\end{table}

\begin{table}[htbp]
\small
\centering
\caption{Normalized-curve CV across LLMs: word dropout stays tight (2.0\%) while every character-level noise type spreads (4.6--5.5\%).}
\label{tab:realistic_noise}
\begin{tabular}{lcc}
\toprule
Noise type & Scale & CV \\
\midrule
Random character & character & 4.8\% \\
Keyboard typo (QWERTY) & character & 5.5\% \\
OCR confusion & character & 4.6\% \\
Adjacent swap & character & 4.9\% \\
\midrule
Word dropout & word & 2.0\% \\
\bottomrule
\end{tabular}
\end{table}

\end{document}